# TOPOLOGICAL DEEP LEARNING FOR SPEECH DATA

ZHIWANG YU

ABSTRACT. Topological data analysis (TDA) offers novel mathematical tools for deep learning. Inspired by Carlsson et al., this study designs topology-aware convolutional kernels that significantly improve speech recognition networks. Theoretically, by investigating orthogonal group actions on kernels, we establish a fiber-bundle decomposition of matrix spaces, enabling new filter generation methods. Practically, our proposed Orthogonal Feature (OF) layer achieves superior performance in phoneme recognition, particularly in low-noise scenarios, while demonstrating cross-domain adaptability. This work reveals TDA's potential in neural network optimization, opening new avenues for mathematics-deep learning interdisciplinary studies.

## 1. Introduction

### 1.1. Backgrounds

This study aims to integrate topological data analysis with deep neural networks, focusing on the application of topological convolutional kernels in speech recognition tasks, particularly for phoneme identification and word classification. By incorporating topological feature extraction mechanisms, we seek to enhance the network's ability to capture key topological characteristics in speech signals, thereby improving recognition performance.

The development of deep neural networks has undergone several critical phases. Early fully connected networks were constrained by computational limitations and theoretical understanding until the emergence of convolutional neural networks (CNNs), which marked the golden age of deep learning. CNNs significantly reduced parameter complexity through local connectivity and weight sharing while preserving essential spatial hierarchical features. However, traditional CNNs exhibit inherent limitations in comprehending the global topological properties of data. In this context, topological neural networks emerged as a promising solution. In 2004 [1], they first identified the topological structure of three rings in data using persistent homology methods; in 2008 [2], they successfully detected the Klein bottle, a complex topological manifold; in 2009 [3], they proposed a theoretical framework explaining the significance of topological features in data analysis; in 2018 [4], they extended topological analysis methods to the study of neural network weight distributions; and in 2023 [5], Love et al. achieved a breakthrough by directly encoding topological features in the design of convolutional kernel. These groundbreaking contributions provide the theoretical basis for our research, particularly in integrating topological feature extraction mechanisms into speech-processing-specific convolutional kernel designs.

Traditional speech processing methods often overlook the rich topological structures embedded in speech signals. The complex patterns formed by speech signals in the time-frequency domain contain crucial topological features that play a vital role in phoneme discrimination and word recognition. By introducing topological data analysis tools, we can more effectively capture these structural characteristics. Notably, the local extrema and connectivity relationships formed by speech signals in the mel-spectrogram constitute specific topological configurations, which exhibit systematic differences across phonemes. The topological convolutional kernel proposed in this study is specifically designed to model these features, enabling simultaneous extraction of conventional spectral features while explicitly representing the topological properties of speech signals. This approach not only improves recognition accuracy but also enhances the model's robustness to noise and variations, offering a new technical pathway for speech recognition systems in complex environments.

## 1.2. Statement of Results

The main focus of this dissertation is the application of convolutional kernels, constructed using the newly defined Orthogonal Feature Layer (OF), to phoneme recognition. Furthermore, the approach is generalized to word recognition and image recognition, demonstrating its versatility and adaptability across multiple domains.

First, in Chapter 3, we consider the space $M_{3\times3}(\mathbb{R})$, which is the most common space of convolutional kernels, as $\{[\boldsymbol{v}_1, \boldsymbol{v}_2, \boldsymbol{v}_3]\}$. Without loss of generality, define the subspace $M = \{[\boldsymbol{v}_1, \boldsymbol{v}_2, \boldsymbol{v}_3] | \boldsymbol{v}_1 + \boldsymbol{v}_2 + \boldsymbol{v}_3 = 0\}$ with the Frobenius form of the matrix in $M$ equal to one. Next, we define a group action on $M$ by

$$\theta(\boldsymbol{Q}, \boldsymbol{m}) = \boldsymbol{Q}\boldsymbol{m} \text{ for } \boldsymbol{Q} \in \text{SO}(3) \text{ and } \boldsymbol{m} \in M.$$

Thus, denote the quotient map from $M$ to $M/\text{SO}(3)$ by $\pi$, then the orbit space $M/\text{SO}(3)$, denoted as $B$, is homeomorphic to a disk $D^2$. Moreover, there exists a stratified fiber bundle on $B$. The fiber is $\text{SO}(3)/(\text{SO}(2)\rtimes\mathbb{Z}_2) \cong \mathbb{R}P^2$ on the case when $\boldsymbol{v}_1 + \boldsymbol{v}_3 = 0$, and $\text{SO}(3)/\text{SO}(2) \cong S^2$ on the case when $\boldsymbol{v}_1$ and $\boldsymbol{v}_3$ are collinear, and $\text{SO}(3)/\mathbb{Z}_2 \cong L(4,1)$ on the case when $\boldsymbol{v}_1$ and $\boldsymbol{v}_3$ are have equal magnitudes, and $\text{SO}(3)$ on the other cases. The above provides a representation of $M$ by special orthogonal group action.

Second, in Chapter 4, we define the Orthogonal Feature Layer (OF) by selecting elements in $B$ and SO(3). Then, we compare neural networks constructed using OF convolutional kernels to traditional neural networks and the networks proposed by Love et al. on phoneme datasets. The results indicate that OF achieves the highest accuracy under low-noise conditions. However, in high-noise environments, OF's performance declines, with KF (kernel filters) emerging as the superior approach.

Finally, in Chapter 5, the applicability of OF convolutional kernels is further explored by extending their use to word datasets and image datasets. Results demonstrate consistent generalization properties that showcase the versatility and robustness of the proposed methodology.

In this article, we treat the weight vector and convolutional kernel interchangeably, without differentiating between the two concepts. The integration of TDA and CNN is presented not only as a promising research direction but also as a practical solution that addresses some of the key challenges in modern data science. For details, all experiments can be seen at (`https://github.com/ZhiwangYu/OrthogonalityFeatures`).

## 1.3. Outline

This dissertation integrates theories, methodologies, and applications of topological deep learning through five systematically organized chapters. Chapter 2 introduces fundamental preliminaries, primarily focusing on Convolutional Neural Networks (CNNs), Topological Convolutional Neural Networks, and phonemes. Chapter 3 formalizes the spectrogram convolutional kernel space through geometric and topological constraints, including contrast maximization and group actions, providing a structured framework for speech processing. This leads to Chapter 4, where novel kernel designs are introduced and validated, showing improved phoneme recognition accuracy and noise robustness. Chapter 5 discusses broader implications, including model performance on unfiltered phonemes, generalization to word and image tasks.

## 2. Preliminaries

### 2.1. Convolutional Neural Network

**Definition 2.1** (CNN Topology). *A Convolutional Neural Network is a feedforward system* $\mathcal{N} = (V, E, \Lambda)$ *where:*

*(1) The vertex set* $V = \bigsqcup_{k=0}^{L} V_k$ *decomposes into layers with* $V_0$ *(input) and* $V_L$ *(output)*

*(2) Edges* $E \subset \bigcup_{k=0}^{L-1}(V_k \times V_{k+1})$ *respect layer ordering*

*(3) Weight parameters* $\Lambda = \{\lambda_e \in \mathbb{R}\}_{e \in E}$ *exhibit translational symmetry*

The CNN dynamics are governed by two fundamental constraints:

**Spatial Locality**: For convolutional layers $V_k = \chi_k \times \mathbb{Z}^d$, each $(v, w) \in E$ satisfies $w = (\kappa', \boldsymbol{x}')$ only if $||\boldsymbol{x} - \boldsymbol{x}'||_\infty \leq r_k$ for some receptive field radius $r_k$.

**Parameter Sharing**: Weight values $\lambda_{(\kappa, \boldsymbol{x}),(\kappa', \boldsymbol{x}')}$ depend solely on $\kappa, \kappa'$ and the displacement $\boldsymbol{x} - \boldsymbol{x}'$.

**Definition 2.2** (Feature Propagation). *The activation* $a_w$ *at node* $w \in V_{k+1}$ *computes as:*

$$a_w = \sigma \left( \sum_{\substack{v \in V_k \\ (v,w) \in E}} \lambda_{(v,w)} a_v + b_w \right)$$

*where* $\sigma$ *denotes the ReLU activation and* $b_w$ *denotes the bias term.*

CNNs implement multiscale processing through interleaved operations:

**Convolutional Blocks**: Combine spatial filtering (via learned kernels) with pointwise nonlinearities. Each block transforms feature maps $F_k : \mathbb{Z}^d \to \mathbb{R}^{c_k}$ to $F_{k+1} : \mathbb{Z}^d \to \mathbb{R}^{c_{k+1}}$ through (where $c_i := \dim(F_i)$ is the channel dimension at layer $i$):

$$F_{k+1}(\boldsymbol{x}) = \sigma \left( \sum_{||\boldsymbol{y}|| \leq r} K(\boldsymbol{y}) F_k(\boldsymbol{x} + \boldsymbol{y}) + \boldsymbol{b} \right),$$

where $\mathbf{x} \in \mathbb{Z}^d$ is spatial position in output feature map, and where $\mathbf{y} \in \{-r, ..., r\}^d$ is offset within convolutional kernel, and where $K(\mathbf{y}) \in \mathbb{R}$ is kernel weight at offset $\mathbf{y}$, and where $F_k(\mathbf{x} + \mathbf{y})$ is input feature at position $\mathbf{x} + \mathbf{y}$, and where $b \in \mathbb{R}$ is bias term, and where $\sigma$ is nonlinear activation (e.g. ReLU), and where $r$ is kernel radius ($3 \times 3$ kernel $\Rightarrow r = 1$).

**Downsampling**: Pooling layers induce spatial compression by local aggregation, typically via max or average operations over $s \times s$ windows.

**Remark 2.3.** *The architectural constraints of CNNs - locality, weight sharing, and hierarchical composition - encode an implicit prior favoring translation-equivariant feature detection while maintaining parametric efficiency.*

## 2.2. Topological Convolutional Neural Network

To contextualize our analytical framework (cf. Carlsson[3]), we adopt a function-theoretic viewpoint of the Klein bottle. The $3 \times 3$ image patches are interpreted as discrete samples obtained by evaluating smooth functions $f : D \to \mathbb{R}$ at nine predetermined grid points $\{p_k\}_{k=1}^{9} \subset D$. Our investigation focuses on identifying closed subspaces $\mathcal{F} \subset C(D, \mathbb{R})$ that satisfy the approximation property:

$$\sup_{f \in \mathcal{F}} ||f||_{L^2(\{p_k\})} \approx ||f||_{L^2(D)},$$

where the left-hand norm corresponds to patch space measurements.

Let $\mathcal{Q}$ denote the space of bivariate quadratic polynomials, explicitly parametrized as

$$f(x,y) = A + Bx + Cy + Dx^2 + Exy + Fy^2 \quad (A, \dots, F \in \mathbb{R}).$$

This constitutes a six-dimensional real vector space. Our analysis focuses on the constrained subspace $\mathcal{P} \subseteq \mathcal{Q}$ defined by the conditions

$$\int_D f(x,y) \, \mathrm{d}x\mathrm{d}y = 0 \quad \text{(mean centering)}, \quad \int_D f(x,y)^2 \, \mathrm{d}x\mathrm{d}y = 1 \quad \text{(contrast normalization)}.$$

The linear constraint alone reduces $\mathcal{Q}$ to a five-dimensional affine subspace, while the quadratic normalization further restricts $\mathcal{P}$ to a four-dimensional ellipsoid embedded within this subspace.

We subsequently characterize the submanifold $\mathcal{P}_0 \subseteq \mathcal{P}$ consisting of functions with the specialized form

$$f(x,y) = q(\lambda x + \mu y),$$

where $q$ is a single-variable quadratic function, and $\lambda^2 + \mu^2 = 1$. The space of such functions within $\mathcal{Q}$ is 4-dimensional-three parameters define $q$, and $(\lambda, \mu)$ lies on the unit circle, which is one-dimensional. Incorporating the two additional constraints reduces this to a 2-dimensional complex $\mathcal{P}_0$.

We demonstrate that $\mathcal{P}_0$ is homeomorphic to the Klein bottle $\mathcal{K}$ via the following construction. Define the function space $A$ as containing all univariate quadratic polynomials of the form

$$q(t) = c_0 + c_1 t + c_2 t^2 \quad (c_i \in \mathbb{R})$$

subject to the integral constraints

$$\int_{-1}^{1} q(t)\mathrm{d}t = 0 \quad \text{(zero mean)}, \quad \int_{-1}^{1} q^2(t)\mathrm{d}t = 1 \quad \text{(unit energy)}.$$

Consider the $\mathcal{K}$ by quotient maps. The original space $\mathcal{Q}$ is homeomorphic to $\mathbb{R}^3 \times S^1$. The mean centering can be considered as a quotient $\theta_1$ as

$$\theta_1(q) = q_1,$$

where $q(t) = c_0 + c_1 t + c_2 t^2$ and $q_1(t) = c_{01} + c_1 t + c_2 t^2$ satisfying the mean centering condition. The unit energy can be considered as a quotient $\theta_2$ as

$$\theta_1(q) = q_2,$$

where $q_2 = \frac{q}{||q||_2}$.

Define the involution $f : \mathcal{Q} \to \mathcal{Q}$ by

$$f(q)(t) = q_0(t) = c_0 - c_1 t + c_2 t^2,$$

which reverses the sign of the linear term $c_1$. This satisfies $f^2 = \mathrm{id}$.

The quotient $\theta_1$ enforces $\int_{S^1} q(t)dt = 0$, eliminating $c_0$. The reduced space is:

$$\theta_1(\mathcal{Q}) \cong \mathbb{R}^2 \times S^1 \quad (\text{parameters } (c_1, c_2) \in \mathbb{R}^2, \ t \in S^1).$$

Under $f$, the coefficients transform as $(c_1, c_2) \mapsto (-c_1, c_2)$.

The quotient $\theta_2$ normalizes the energy:

$$\theta_2(q) = \frac{(c_1, c_2)}{||(c_1, c_2)||_2} \in S^1 \quad (\text{unit circle}).$$

The resulting space after $\theta_2$ is a fiber bundle over $S^1$ with fiber $S^1$.

The involution $f$ acts on the normalized coefficients as:

$$f : (c_1, c_2) \mapsto (-c_1, c_2) \Longrightarrow (\cos\theta, \sin\theta) \mapsto (\cos(\pi - \theta), \sin(\pi - \theta)).$$

This corresponds to a reflection $\theta \mapsto \pi - \theta$ on $S^1$. Simultaneously, the base $S^1$ (original $t \in S^1$) is twisted by a half-period shift $t \mapsto t + \pi$ due to the phase dependency in $\mathcal{Q}$. The total space is constructed by gluing the fibers $S^1$ over the base $S^1$ with a reflection map. This gluing is equivalent to the Klein bottle:

$$\mathcal{K} \cong (S^1 \times S^1)/\sim, \quad (\theta, t) \sim (\pi - \theta, t + \pi).$$

Since the involution $f$ introduces a non-orientable twist in both the fiber and base, the quotient space is the Klein bottle.

## 2.3. Phoneme

### 2.3.1. Phonetic Building Blocks

Phonemes are systematically classified into vowels and consonants according to articulatory characteristics. The rhythmic interplay between these units forms the structural basis of spoken language. This hierarchical organization drives research emphasis toward suprasegmental analysis (words/sentences), where expanded contextual dependencies enable more reliable pattern identification.

Modern systems employ a three-tiered processing hierarchy:

- **Phoneme Level**: 40-60 basic units (English: 44 phonemes) with 50-200ms duration
- **Syllable Level**: 10,000+ possible combinations through phoneme concatenation
- **Prosodic Level**: Pitch contours and stress patterns conveying semantics

The precise alignment between transient acoustic features and discrete phonetic symbols remains challenging, particularly for coarticulated phonemes where adjacent sounds blend spectrally.

### 2.3.2. Phonetic Classification via IPA Standards

The International Phonetic Alphabet (IPA) categorizes phonemes into three primary classes: pulmonic consonants, non-pulmonic consonants, and vowels. Our analysis focuses exclusively on pulmonic consonants and vowels, as non-pulmonic consonants exhibit negligible prevalence in English. Pulmonic consonants are produced by constricting airflow at the glottis (the space between vocal folds) or oral cavity while coordinating pulmonary airflow, exemplified by symbols such as [b], [p], [m], and [n].

Consonants are further specified through three articulatory dimensions:

(1) **Place**: Bilabial [p], Alveolar [t], Velar [k]
(2) **Manner**:
    - Plosives [ptk]

- Fricatives [szf]
- Nasals [mnŋ]
- Approximants [jw]

(3) **Voicing**: Vocal fold vibration (e.g., [z] vs. [s])

Vowels are systematically mapped in IPA based on lingual positioning (Figure 1), quantified through:

- **Height**:
    - High [i]
    - Mid [e]
    - Low [a]
- **Backness**:
    - Front [i]
    - Central [ə]
    - Back [u]
- **Roundedness**:
    - Rounded [y]
    - Unrounded [i]

To streamline English phonetic notation, ARPABET emerged as a practical alternative, mapping 39 English phonemes to ASCII combinations (Figure 2). This system enables efficient computational processing through:

- Single-letter vowels: AA [ɑ], AE [ae]
- Two-letter consonants: SH [ʃ], TH [θ]
- Stress markers: Primary (ˈ), secondary (ˌ)

The above two figures (see Figure 1 and Figure 2) are both from Wikipidea (`https://en.wikipedia.org/wiki/International_Phonetic_Alphabet`; `https://en.wikipedia.org/wiki/ARPABET`).

### 2.3.3. STFT and Spectrogram

The conversion of raw speech waveforms into spectrograms begins with the Short-Time Fourier Transform (STFT), which decomposes the signal into its frequency components across time intervals [6].

Formally, the STFT of a signal $x(t)$ is given by:

$$X(f,t) = \int_{-\infty}^{\infty} x(\tau)w(\tau - t)e^{-j2\pi f\tau}\,\mathrm{d}\tau,$$

where $w(t)$ denotes a window function (such as Hamming or Gaussian windows) centered at each temporal point $t$, and $f$ corresponds to the frequency domain. This approach captures localized frequency content while preserving temporal resolution.

To illustrate the transformation of speech signals from waveforms to spectrograms, we apply the Short-Time Fourier Transform (STFT). This process captures temporal and frequency-domain features, providing a foundation for subsequent audio analysis. The following figure demonstrates an example of a speech waveform (top) and its corresponding spectrogram (bottom), offering a clear visualization of how sound evolves across time and frequency domains.

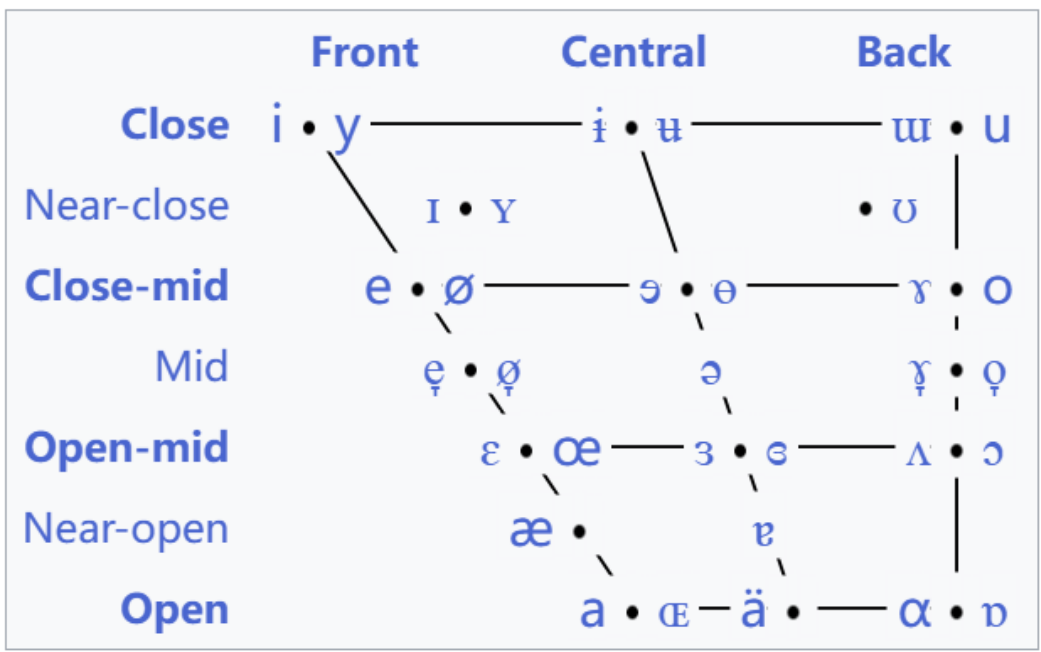

Figure 1. Positioning of vowels in oral cavity

| | | |
|---|---|---|
| AA | ɑ~ɒ | balm, bot (with father–bother merger) |
| AE | æ | bat |
| AH | ʌ | butt |
| AO | ɔ | caught, story |
| AW | aʊ | bout |
| AX | ə | comma |
| AXR[1] | ɚ | letter, forward |
| AY | aɪ | bite |
| EH | ɛ | bet |
| ER | ɝ | bird, foreword |
| EY | eɪ | bait |
| IH | ɪ | bit |
| IX | ɨ | roses, rabbit |
| IY | i | beat |
| OW | oʊ | boat |
| OY | ɔɪ | boy |
| UH | ʊ | book |
| UW | u | boot |
| UX[1] | ʉ | dude |

(a) ARPABET vowel-phoneme mapping table

| | | |
|---|---|---|
| B | b | buy |
| CH | tʃ | China |
| D | d | die |
| DH | ð | thy |
| DX | ɾ | butter |
| EL | l̩ | bottle |
| EM | m̩ | rhythm |
| EN | n̩ | button |
| F | f | fight |
| G | g | guy |
| HH or H[1] | h | high |
| JH | dʒ | jive |
| K | k | kite |
| L | l | lie |
| M | m | my |
| N | n | nigh |
| NX or NG[2] | ŋ | sing |
| NX[1] | ɾ̃ | winner |

(b) ARPABET consonant mapping table I

| | | |
|---|---|---|
| P | p | pie |
| Q | ʔ | uh-oh |
| R | ɹ | rye |
| S | s | sigh |
| SH | ʃ | shy |
| T | t | tie |
| TH | θ | thigh |
| V | v | vie |
| W | w | wise |
| WH | ʍ | why (without wine–whine merger) |
| Y | j | yacht |
| Z | z | zoo |
| ZH | ʒ | pleasure |

(c) ARPABET consonant mapping table II

Figure 2. ARPABET phonetic notation system

## 3. The Space of Spectrogram Convolution Kernels

In this chapter, we consider the group action of the third-order special orthogonal group on the space of $3 \times 3$ real matrices. By leveraging the invariance properties of the group action, we first reduce the dimensionality of the matrix space to five. Subsequently, a new representation of the matrix space is introduced through orbit spaces and the special orthogonal group.

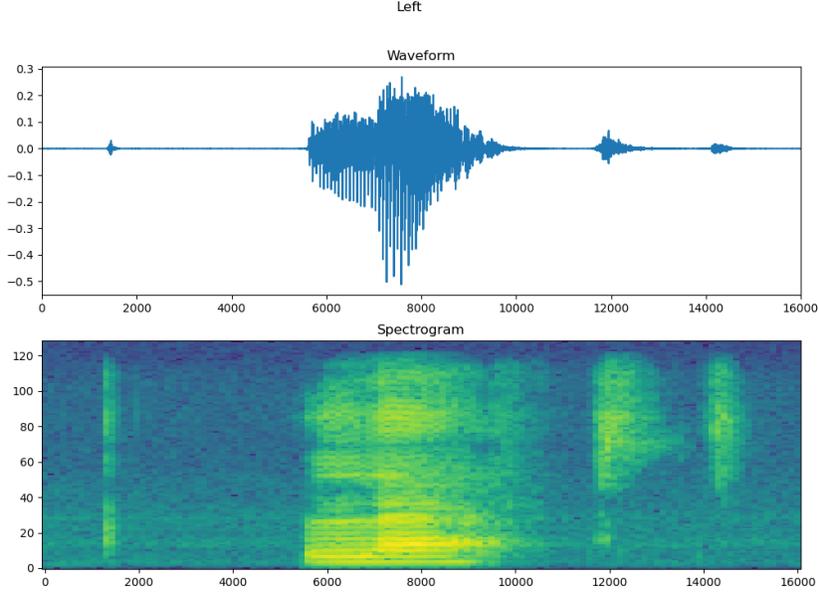

FIGURE 3. Waveform and Corresponding Spectrogram on mini speech commands

### 3.1. The Space of High-Contrast Spectrogram Convolution Kernels

Spectrograms, unlike ordinary images, lose their semantic interpretation under rotation. Thus, when considering convolution kernels for spectrograms, which we interpret as local fragments of speech where the variation is predominantly along the temporal axis, it is natural to restrict our attention to kernels that reflect this asymmetry.

**Definition 3.1** (Kernel Norm). *Let*

$$\boldsymbol{A} = [\boldsymbol{v}_1, \boldsymbol{v}_2, \boldsymbol{v}_3] \in M_{3\times 3}(\mathbb{R})$$

*be a $3 \times 3$ convolution kernel with column vectors $\boldsymbol{v}_1, \boldsymbol{v}_2, \boldsymbol{v}_3 \in \mathbb{R}^3$. Define the norm of $\boldsymbol{A}$ by*

$$||\boldsymbol{A}|| = \sqrt{||\boldsymbol{v}_1||^2 + ||\boldsymbol{v}_2||^2 + ||\boldsymbol{v}_3||^2}.$$

*Note that this norm is equivalent to the standard $L^2$-norm up to a constant factor.*

**Definition 3.2** (Contrast). *The **contrast** of a convolution kernel $\boldsymbol{A} = [\boldsymbol{v}_1, \boldsymbol{v}_2, \boldsymbol{v}_3] \in M$ is defined by*

$$\mathrm{con}(\boldsymbol{A}) = \sqrt{||\boldsymbol{v}_1 - \boldsymbol{v}_2||^2 + ||\boldsymbol{v}_2 - \boldsymbol{v}_3||^2}.$$

**Remark 3.3.** *The use of the contrast measure is motivated by the observation that spectrograms are inherently directional. Since rotation typically destroys the temporal structure of a* *spectrogram, a high-contrast convolution kernel (with respect to the temporal axis) is desirable for effectively capturing local speech features.*

We now introduce a constrained space of convolution kernels that are both normalized and optimized for high contrast.

**Definition 3.4** (Normalized Convolution Kernels)**.** *We consider the subspace of $M_{3\times3}(\mathbb{R})$ consisting of convolution kernels $\boldsymbol{A} = [\boldsymbol{v}_1, \boldsymbol{v}_2, \boldsymbol{v}_3]$ satisfying the unit norm condition*

$$||\boldsymbol{A}|| = 1.$$

**Definition 3.5** (Contrast-Maximizing Constraint)**.** *In order to maximize contrast, we further impose the constraint that the kernel belongs to the orthogonal complement of the zero-contrast subspace. Concretely, we require*

$$\boldsymbol{v}_1 + \boldsymbol{v}_2 + \boldsymbol{v}_3 = 0.$$

**Definition 3.6** (The Kernel Space $M$)**.** *Let $M$ denote the set of all $3 \times 3$ convolution kernels satisfying*

$$||\boldsymbol{A}|| = 1 \quad and \quad \boldsymbol{v}_1 + \boldsymbol{v}_2 + \boldsymbol{v}_3 = 0.$$

*Then,*

$$M = \{\, \boldsymbol{A} \in M_{3\times3}(\mathbb{R}) \mid ||\boldsymbol{A}|| = 1,\ \boldsymbol{v}_1 + \boldsymbol{v}_2 + \boldsymbol{v}_3 = 0 \,\}.$$

**Theorem 3.7.** *The space $M$ is homeomorphic to the $5$-dimensional sphere $S^5$.*

*Proof.* [Sketch of Proof] The constraints $||\boldsymbol{A}|| = 1$ and $\boldsymbol{v}_1 + \boldsymbol{v}_2 + \boldsymbol{v}_3 = 0$ define a smooth submanifold of $M_{3\times3}(\mathbb{R})$. One may show via dimension counting and the implicit function theorem that this submanifold has dimension $9 - 3 - 1 = 5$ (since $M_{3\times3}(\mathbb{R}) \cong \mathbb{R}^9$, and the two constraints remove 4 degrees of freedom). An explicit construction or application of known results then shows that this 5-dimensional manifold is in fact diffeomorphic to (and hence homeomorphic to) the standard sphere $S^5$. $\square$

## 3.2. Group Action and Quotient Space

Since $M \subset M_{3\times3}(\mathbb{R})$, the group of orthogonal transformations acts naturally on $M$.

**Definition 3.8** (Orthogonal Group Action)**.** *Let $\theta : \mathrm{SO}(3) \times M \to M$ be defined by*

$$\theta(\boldsymbol{Q}, \boldsymbol{m}) = \boldsymbol{Q}\boldsymbol{m}, \quad for\ \boldsymbol{Q} \in \mathrm{SO}(3)\ and\ \boldsymbol{m} \in M.$$

*Then $\theta$ is a smooth group action.*

**Theorem 3.9** (Contrast Projection for General Matrices)**.** *For any matrix $\boldsymbol{A} \in M_{3\times3}(\mathbb{R}^3)$ except when the three column vectors are identical, in which case the contrast is defined as $0$, such as the matrix $\begin{bmatrix} 1 & 1 & 1 \\ 0 & 0 & 0 \\ -1 & -1 & -1 \end{bmatrix}$, the following procedure projects it onto the constrained subspace $M$:*

*(1)* ***Orthogonal Transformation****: Apply an orthogonal matrix $\boldsymbol{Q} \in \mathrm{SO}(3)$ to transform the sum of column vectors into a uniform vector:*

$$(1) \qquad \boldsymbol{Q}(\boldsymbol{v}_1 + \boldsymbol{v}_2 + \boldsymbol{v}_3) = \lambda\mathbf{1}, \quad \lambda \in \mathbb{R},\ \mathbf{1} = (1, 1, 1)^\top.$$

*(2)* **Centering**: *Subtract the mean value from each component:*

$$\tilde{\boldsymbol{A}} = \boldsymbol{Q}\boldsymbol{A} - \frac{\lambda}{3}\mathbb{1}\mathbb{1}^{\top}$$

*The resulting matrix $\tilde{\boldsymbol{A}}$ satisfies $\tilde{\boldsymbol{v}}_1 + \tilde{\boldsymbol{v}}_2 + \tilde{\boldsymbol{v}}_3 = 0$, i.e., $\tilde{\boldsymbol{A}} \in M$.*

*Proof.* [Rationale] This projection ensures:

- Invariance under orthogonal transformations: $\|\boldsymbol{Q}\boldsymbol{v}\| = \|\boldsymbol{v}\|$.
- Translation invariance: $\boldsymbol{v}_i \mapsto \boldsymbol{v}_i + \boldsymbol{c}$ cancels in (2).

The contrast $\mathrm{con}(\tilde{\boldsymbol{A}}) = \sqrt{\|\tilde{\boldsymbol{v}}_1 - \tilde{\boldsymbol{v}}_2\|^2 + \|\tilde{\boldsymbol{v}}_2 - \tilde{\boldsymbol{v}}_3\|^2}$ on $M$ inherits these properties.    □

Note that for any $\boldsymbol{m} \in M$, and for any $\boldsymbol{Q} \in \mathrm{SO}(3)$, the group action defined above is compatible with the previously defined contrast, that is,

$$\mathrm{con}(\boldsymbol{Q}\boldsymbol{m}) = \mathrm{con}(\boldsymbol{m}).$$

**Definition 3.10** (Quotient Space $B$). *Define the homogeneous space (or orbit space)*

$$B = M/\mathrm{SO}(3),$$

*i.e., two kernels in $M$ are identified if one can be obtained from the other by an orthogonal transformation.*

Given coordinates derived from the columns of a kernel, let

$$x = \|\boldsymbol{v}_1\|^2, \quad y = \|\boldsymbol{v}_3\|^2, \quad z = \boldsymbol{v}_1 \cdot \boldsymbol{v}_3.$$

Then the constraints in $M$ imply the following relations:

$$x + y + z = \frac{1}{2},$$
$$z^2 \leq xy.$$

**Proposition 3.11.** *The quotient space $B = M/\mathrm{SO}(3)$ is homeomorphic to the closed disk $D^2$.*

*Proof.* [Sketch of Proof] By introducing the coordinates $(x, y, z)$ and considering the relations

$$x + y + z = \frac{1}{2} \quad \text{and} \quad z^2 \leq xy,$$

one can show that the set of equivalence classes is parametrized by two independent parameters satisfying inequalities that define a closed 2-dimensional disk. A more detailed study of the invariants associated with the $\mathrm{SO}(3)$-action yields the claim that $B$ is homeomorphic to $D^2$.    □

In particular, the boundary of $B$, called $\partial B$, corresponds to the case where the equality $z^2 = xy$ holds in the relation (3).

**Remark 3.12.** *For an element $\left[ \frac{1}{\sqrt{6}} \begin{bmatrix} 1 & 0 & -1 \\ 1 & 0 & -1 \\ 1 & 0 & -1 \end{bmatrix} \right] \in B$, the preimage set in $M$ is*

$$\left\{ \begin{bmatrix} a & 0 & -a \\ b & 0 & -b \\ c & 0 & -c \end{bmatrix} \middle| a^2 + b^2 + c^2 = \frac{1}{2} \right\}.$$

*However, the dimension of preimage is two, not three, which shows that $M \to B$ is not a fiber bundle.*

Fortunately, there exists a stratified fiber bundle over $B$. Specifically:

- On the boundary of $B$, denoted $\partial B$, the fiber is $\mathrm{SO}(3)/\mathrm{SO}(2) \cong S^2$, quotienting out rotations around a fixed axis.
- On the region where $x = y$, the fiber is $\mathrm{SO}(3)/\mathbb{Z}_2 \cong L(4, 1)$, quotienting out rotations by $180°$ about a fixed axis. Here $L(4, 1)$ is a Lens space.
- On the intersection of the two aforementioned cases, i.e. $\boldsymbol{v}_1 + \boldsymbol{v}_3 = 0$, the fiber is given by $\mathrm{SO}(3)/(\mathrm{SO}(2) \rtimes \mathbb{Z}_2)$, which is isomorphic to $\mathbb{R}P^2$ (the real projective plane).
- On the remaining portion, the structure forms a principal $\mathrm{SO}(3)$-bundle.

**Proposition 3.13.** *For any $(x, y)$ satisfying the relations (3), one corresponding kernel can be selected by $\begin{bmatrix} \boldsymbol{v}_1 & -\boldsymbol{v}_1 - \boldsymbol{v}_2 & \boldsymbol{v}_2 \end{bmatrix}$, where $\boldsymbol{v}_1 = \sqrt{\frac{x}{3}} \begin{bmatrix} 1 \\ 1 \\ 1 \end{bmatrix}$ and $\boldsymbol{v}_2 = \sqrt{\frac{y}{3}} \cos\phi \begin{bmatrix} 1 \\ 1 \\ 1 \end{bmatrix} + \sqrt{\frac{y}{6}} \sin\phi \begin{bmatrix} 1 \\ -2 \\ 1 \end{bmatrix}$, where $\sin\phi = \sqrt{1 - \cos^2\phi}$ and $\cos\phi = \frac{\frac{1}{2} - x - y}{\sqrt{xy}}$ for $x, y \neq 0$. In particular, $y = \frac{1}{2}$ for $x = 0$ and $x = \frac{1}{2}$ for $y = 0$.*

**Remark 3.14.** *$(x, y)$ satisfying the relations (3) if and only if*

$$9\left(x + y - \frac{2}{3}\right)^2 + 3(x - y)^2 \leq 1.$$

## 3.3. Summary

To summarize, we have defined a notion of contrast for spectrogram convolution kernels and introduced rigid constraints (unit norm and zero-sum of column vectors) to define a space $M$ of kernels that are well-suited for processing spectrograms. We have established that $M$ is homeomorphic to $S^5$ and that the natural $\mathrm{SO}(3)$-action on $M$ induces a quotient space $B$ that is homeomorphic to a disk $D^2$. These results lay the foundation for further analysis and applications in spectrogram-based speech processing.

## 4. New Spectrogram Convolution Filters

As in Lee et al.[7], there exist 8 basic vectors in the image patch. However, up to constant factors, they will be reduced to just two, since two of them are of zero contrast, and the other of them can be reduced to two vectors through group actions.

We consider taking the orbit space of these two vectors under group actions as convolution kernels, i.e.,

$$\boldsymbol{A}_1 = \boldsymbol{Q} \begin{bmatrix} 1 & 0 & -1 \\ 1 & 0 & -1 \\ 1 & 0 & -1 \end{bmatrix} / \sqrt{6}, \text{ and } \boldsymbol{A}_2 = \boldsymbol{Q} \begin{bmatrix} 1 & -2 & 1 \\ 1 & -2 & 1 \\ 1 & -2 & 1 \end{bmatrix} / \sqrt{18}.$$

Additionally, this chapter focuses exclusively on phoneme-level recognition, as also mentioned in Chapter 4. Regarding the dataset, we cannot directly obtain phoneme-level annotations but instead employ segmentation tools. The Montreal Forced Aligner (MFA) tool from the SpeechBox dataset[8] is utilized in this study. All segmented phonemes undergo appropriate merging processes: stress variations are not differentiated and are combined, open/close vowel distinctions are eliminated, and highly similar vowel variants are merged. Notably, post-segmentation analysis revealed that certain phonemes with extremely low frequencies tend to be overlooked in prediction models, while overrepresented phonemes create prediction biases. Therefore, all experiments in this chapter employ a balanced subset of 500 samples per phoneme class for classification tasks. Finally, the primary datasets used in this chapter are derived from the SpeechBox corpus, TIMIT[9] and LJSpeech[10] with specific implementation details provided in the experimental section. We selected only half of the LJSpeech dataset because the complete dataset contains a large number of speech signals, which exceeds the processing capability of my computer.

The overall procedure for all experiments in this chapter is as follows: First, segment the audio from the dataset into phonemes through **STFT spectrograms** (Short-Time Fourier Transform). Subsequently, convert the audio corresponding to each phoneme into spectrograms. These spectrograms are then fed into a convolutional neural network (CNN) for training, where the network architecture contains two convolutional layers with 64 filters each, ultimately yielding the classification accuracy.

### 4.1. Orthogonal Feature Layer Construction

Given two initial matrices $\{\boldsymbol{A}_1, \boldsymbol{A}_2\} \in M_{3 \times 3}(R)$, the layer is constructed through the following mathematical operations:

#### 4.1.1. Matrix Augmentation

Extend the matrix set to ensure algebraic closure under inversion:

$$\mathcal{M} = \{\boldsymbol{A}_1, \boldsymbol{A}_2, -\boldsymbol{A}_1, -\boldsymbol{A}_2\}.$$

#### 4.1.2. SO(3)-Informed Kernel Generation

Let $\mathfrak{so}(3)$ denote the Lie algebra with basis generators:

$$\boldsymbol{L}_x = \begin{pmatrix} 0 & 0 & 0 \\ 0 & 0 & -1 \\ 0 & 1 & 0 \end{pmatrix}, \ \boldsymbol{L}_y = \begin{pmatrix} 0 & 0 & 1 \\ 0 & 0 & 0 \\ -1 & 0 & 0 \end{pmatrix}, \ \boldsymbol{L}_z = \begin{pmatrix} 0 & -1 & 0 \\ 1 & 0 & 0 \\ 0 & 0 & 0 \end{pmatrix}.$$

For stochastic kernel generation:

(1) Sample $\theta_x, \theta_y, \theta_z \sim \mathcal{N}(0, \sigma^2)$ independently.

(2) Construct Lie algebra element:

$$\boldsymbol{\theta} = \sum_{i=x,y,z} \theta_i \boldsymbol{L}_i \in \mathfrak{so}(3).$$

(3) Apply exponential map:

$$\boldsymbol{R} = \exp(\boldsymbol{\theta}) \in \mathrm{SO}(3), \quad \text{where } \exp(\boldsymbol{\theta}) = \boldsymbol{I} + \frac{\sin \|\boldsymbol{\theta}\|}{\|\boldsymbol{\theta}\|} \boldsymbol{\theta} + \frac{1 - \cos \|\boldsymbol{\theta}\|}{\|\boldsymbol{\theta}\|^2} \boldsymbol{\theta}^2,$$

where $\|\boldsymbol{\theta}\| = \sqrt{\sum_{i=x,y,z} \theta_i^2}$.

#### 4.1.3. Convolutional Layer Definition

**Definition 4.1** (Orthogonal Features(OF) Convolutional Layer). *Given the kernel space $M$, each conlutional kernel of untrained **Orthogonal Features Convolutional Layer** can be defined by*

$$\boldsymbol{W}_k = \alpha \cdot \boldsymbol{R}_k \boldsymbol{M}_k, \quad \begin{cases} \boldsymbol{R}_k \in \mathrm{SO}(3), \\ \boldsymbol{M}_k \in \mathcal{M}, \\ \alpha \in \mathbb{R}^+ \ (\textit{adjustable scaling factor}). \end{cases}$$

## 4.2. Experimental Result I

In this section, we conducted experiments using the Speechbox dataset to compare the newly proposed OF convolutional layers with multiple other convolutional neural network architectures. The comparative results are presented in the accompanying Figure 4.

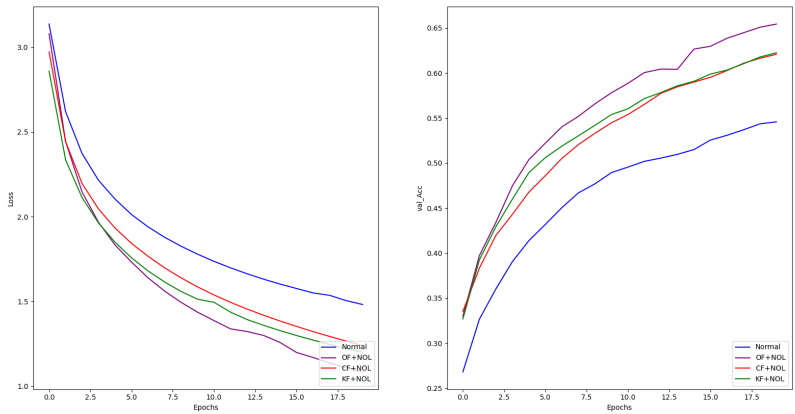

FIGURE 4. Comparisons of Loss and Accuracy on SpeechBox

The comparative analysis reveals two key observations. First, both KF and CF models demonstrate significantly superior performance in phoneme-balanced segmentation compared to traditional CNNs when evaluated against word-level phoneme frequency distributions. Second, and more critically, the proposed OF architecture exhibits marginally better effectiveness than both KF and CF configurations in these phoneme-aware classification tasks.

## 4.3. Experimental Result II

However, if we relax the orthogonality condition to the zero-contrast space, we can obtain a more canonical set of convolution kernels

$$(4) \qquad \boldsymbol{Q} \begin{bmatrix} 1 & 0 & -1 \\ 1 & 0 & -1 \\ 1 & 0 & -1 \end{bmatrix} / \sqrt{6}, \text{ and } \boldsymbol{Q} \begin{bmatrix} 1 & 0 & 1 \\ 1 & 0 & 1 \\ 1 & 0 & 1 \end{bmatrix} / \sqrt{6}.$$

In essence, this set of convolution kernels corresponds to vertical stripe detectors with the middle column set to zero, structured as $[\boldsymbol{v}_1, 0, \pm \boldsymbol{v}_1]$, with which the sphere shares a homeomorphism. For simplicity, the neural network architectures constructed using this set of convolution kernels will retain the nomenclature OF convolutional layers.

First, let us analyze the performance of these convolutional kernel space on the Speechbox dataset (see Figure 5).

Here, we observe that the accuracy has approached 70%, outperforming both the previous orthogonal components and other comparative models.

Experimental results on the two additional datasets, TIMIT and LJSpeech, are also reported, yielding consistent findings (see Figure 6, Figure 7).

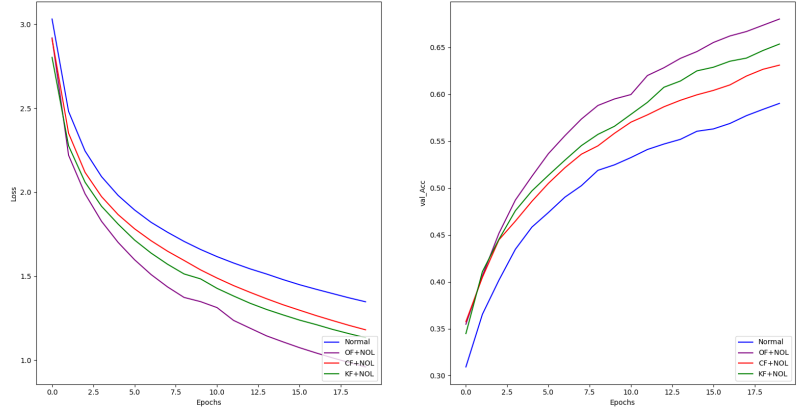

Figure 5. Comparisons of Loss and Accuracy on SpeechBox(Non-Orthogonal)

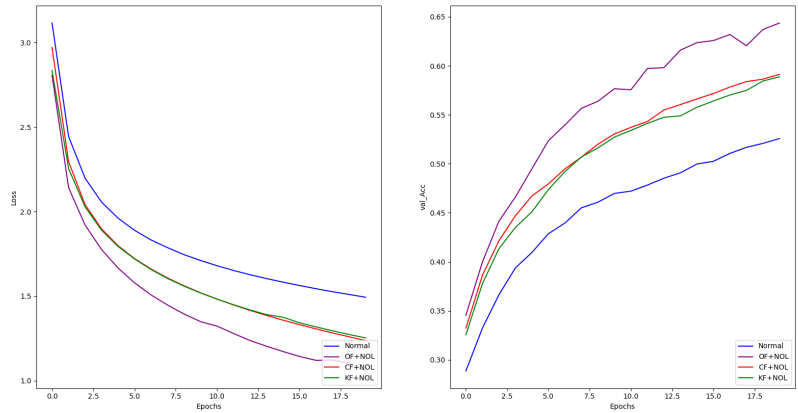

Figure 6. Comparisons of Loss and Accuracy on TIMIT(Non-Orthogonal)

### 4.4. Noise

Analysis of the figure reveals that the datasets exhibit descending accuracy rankings: LJSpeech > SpeechBox > TIMIT, which is likely attributed to variations in acoustic clarity across the datasets. This section investigates the impact of introducing additive white Gaussian noise (AWGN) on model performance.

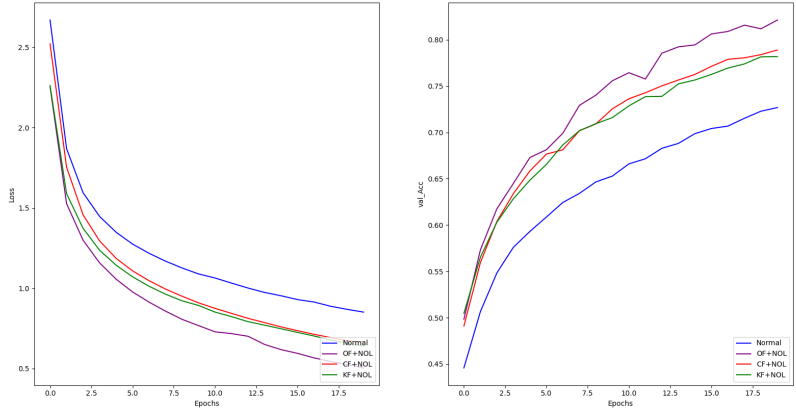

FIGURE 7.  Comparisons of Loss and Accuracy on LJSpeech(Non-Orthogonal)

The additive white Gaussian noise (AWGN) is systematically introduced under controlled signal-to-noise ratio (SNR) conditions, where SNR is mathematically expressed as:

$$\mathrm{SNR}\ (\mathrm{dB}) = 10\log_{10}\left(P_{\mathrm{signal}}/P_{\mathrm{noise}}\right)$$

with $P_{\mathrm{signal}}$ and $P_{\mathrm{noise}}$ representing the power of the original speech signal and the injected Gaussian noise, respectively. The implementation protocol comprises three phases:

(1) Data Partitioning: Split the speech corpus into training and validation subsets.
(2) Noise Injection: Apply AWGN exclusively to the training set across SNR levels ranging from **0 dB** to **20 dB**.
(3) Feature Extraction: Convert the noise-augmented training data into STFT spectrograms for downstream processing, while the validation set remains unaltered to preserve evaluation integrity.

Experimental results on the SpeechBox dataset under varying SNR conditions are as follows (see Figure 8, Figure 9).

The graphical comparison between the aforementioned diagrams demonstrates congruence between the SNR= 20 measurements and their noise-free counterparts. When SNR= 0, OF demonstrates moderate performance, CF exhibits inferior results, whereas KF achieves the optimal performance. This phenomenon might arise from the severe degradation of vertical stripe structures caused by additive noise, leading to reduced accuracy. Consequently, in anti-noise experiments, KF manifests enhanced stability, while OF maintains superior accuracy under low-noise scenarios.

As for the convolutional kernel corresponding to this orthogonal group action, there exist multiple generation approaches, which we omit further elaboration here. In practice, our experiments with several such methods revealed accuracy rates nearly identical to those of the OF+NOL configuration across all aforementioned experimental groups.

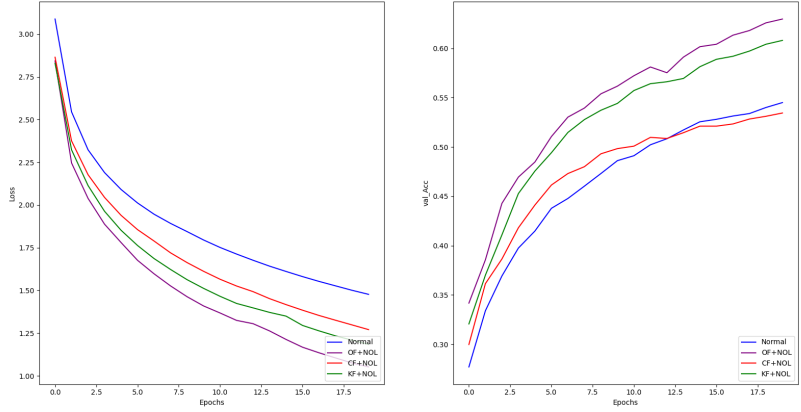

Figure 8. Comparisons of Loss and Accuracy on SpeechBox(SNR= 20)

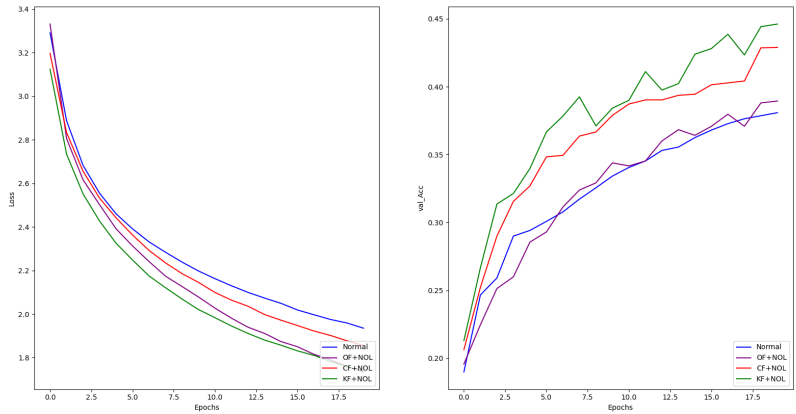

Figure 9. Comparisons of Loss and Accuracy on SpeechBox(SNR= 0)

## 5. Further Applications and Extensions of Theoretical Frameworks

This chapter focuses on addressing gaps and extending prior experimental findings. It begins by supplementing earlier experiments with an analysis of scenarios where no phoneme filtering is applied, providing insights into performance under realistic conditions. Subsequently, it examines how different convolutional neural network architectures perform in word and image classification tasks, showcasing their versatility and efficiency across domains. The discussion then progresses to theoretical advancements, where the analysis of convolutional kernels is extended into the framework of Riemannian geometry, offering a novel perspective on optimization and robustness. Finally, the chapter concludes with an in-depth review of the study's limitations, acknowledging constraints in scope and methodology while outlining directions for future improvement. The convolutional neural network architecture discussed in this chapter is identical to the one in Chapter 6, consisting of two convolutional layers with 64 filters each.

## 5.1. Supplements on Phonemes

While previous noise robustness evaluations were conducted under phoneme-averaged conditions, an idealized scenario deviating from empirical requirements, this section implements dataset-averaged noise testing (without phoneme-level data filtering) to assess performance under more realistic conditions.

The following four figures illustrate the training performance of various neural network architectures across four datasets, SpeechBox (Figure 10), SpeechBox (SNR=0) (Figure 11), TIMIT (Figure 12), and LJSpeech (Figure 13), under conditions where no phoneme count filtering was applied.

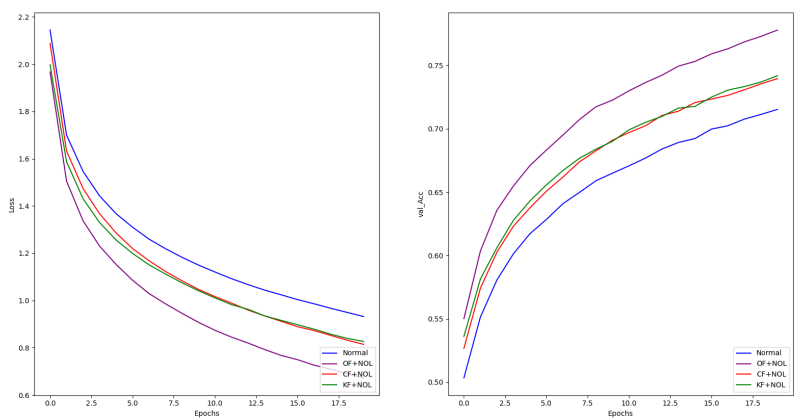

FIGURE 10. Comparisons of Loss and Accuracy on SpeechBox without Selection.

The experimental results align with expectations in that our proposed convolutional kernel remains optimal, particularly under noise-free conditions. However, it is noteworthy that neural networks incorporating circular features and Klein features unexpectedly outperformed traditional architectures, despite prior assertions of their incompatibility with audio tasks. This apparent contradiction may stem from an overlooked preprocessing step: audio normalization was omitted in earlier implementations. Upon revisiting the codebase, we identified this omission as a plausible root cause for the previously observed accuracy degradation.

## 5.2. Applications to Words

Notably, the proposed convolutional layer demonstrates cross-linguistic efficacy, achieving excellent recognition accuracy not only for phoneme-level tasks but also in word-level classification. To systematically validate this capability, this section utilizes the full Speech Commands benchmark dataset[11], a dedicated word-level corpus explicitly designed with approximately balanced frequency distributions across all lexical entries, for comprehensive evaluation.

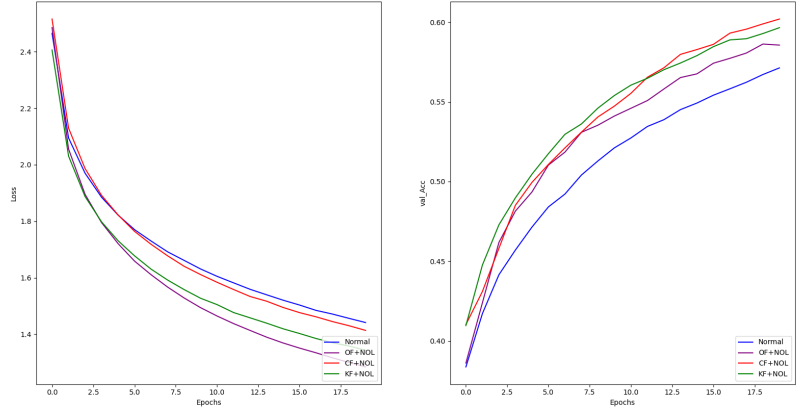

FIGURE 11. Comparisons of Loss and Accuracy on SpeechBox(SNR=0) without Selection.

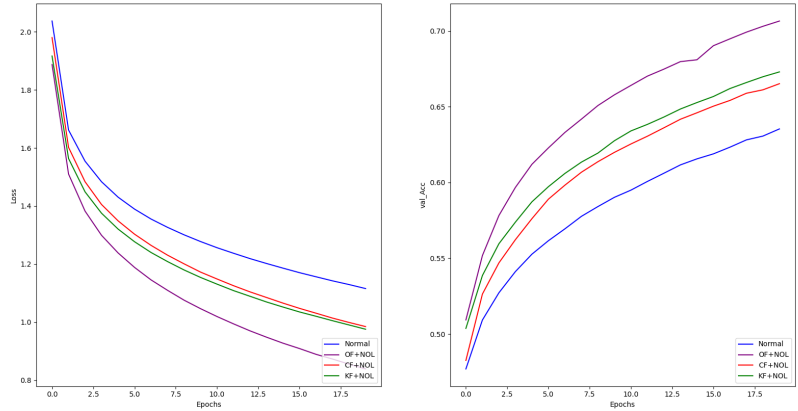

FIGURE 12. Comparisons of Loss and Accuracy on TIMIT without Selection.

Figure 14 demonstrates that our neural network model exhibits robust adaptability to word-level tasks, further validating its versatility across lexical processing challenges.

## 5.3. Applications to Images

Applying these findings retroactively to image processing tasks demonstrates performance metrics comparable to those achieved with Klein bottle configurations, validating the cross-domain adaptability of method. We selected the CIFAR10 dataset [12] for its higher complexity relative to MNIST, providing a more challenging benchmark to evaluate model robustness in handling intricate feature representations (see Figure 15).

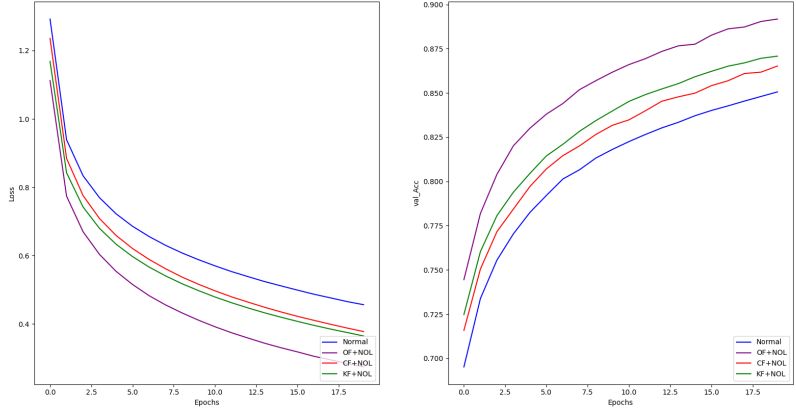

Figure 13. Comparisons of Loss and Accuracy on LJSpeech without Selection.

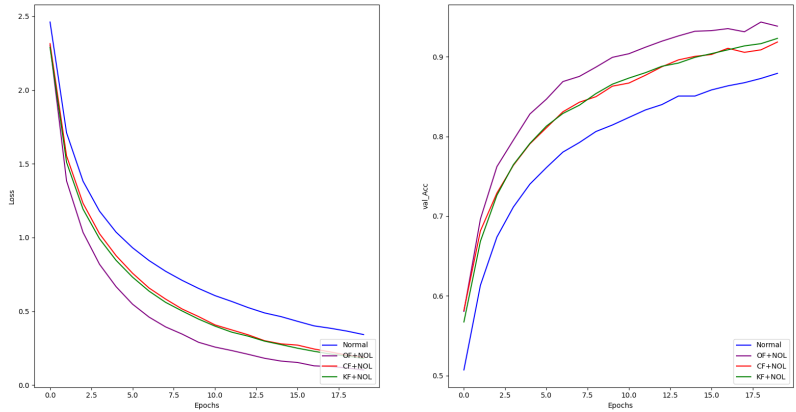

Figure 14. Comparisons of Loss and Accuracy on SpeechCommands.

Figure 15 demonstrates that our model achieves superior performance over conventional neural networks on image-based tasks, while maintaining parity with architectures utilizing Klein features, underscoring its cross-modal versatility.

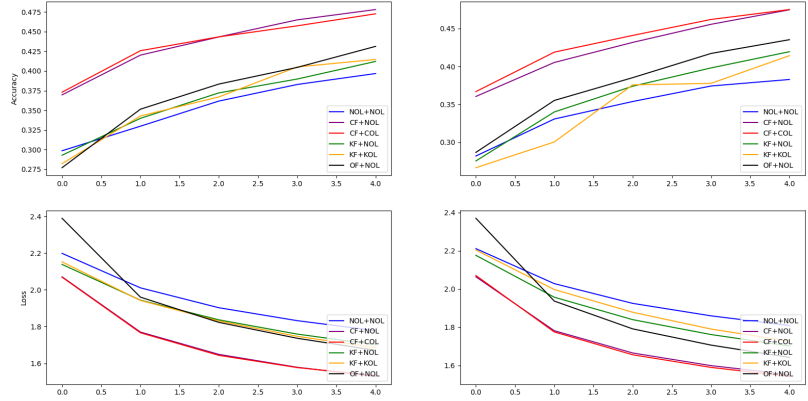

Figure 15. Comparisons of Loss and Accuracy on CIFAR10.

## Acknowledgements

We extend our sincere gratitude to Gunnar Carlsson for his insightful discussions on topological methods in image analysis. We are deeply indebted to Professor Yifei Zhu for his invaluable guidance on the convolutional neural network architecture and its applications to speech processing. Special thanks also go to Qingrui Qu for his instrumental contributions to the implementation and code optimization.

Department of Mathematics, Southern University of Science and Technology, China
*Email address*: `12131239@mail.sustech.edu.cn`